\documentclass[manuscript,nonacm]{acmart}

\usepackage[utf8]{inputenc} 
\usepackage[T1]{fontenc}    
\usepackage{hyperref}       
\usepackage{url}            
\usepackage{booktabs}       
\usepackage{amsfonts}       
\usepackage{nicefrac}       
\usepackage{microtype}      
\usepackage{xcolor}         
\usepackage{amsmath}
\usepackage{booktabs}
\usepackage{cleveref}
\usepackage{graphicx}
\usepackage{placeins} 
\usepackage{xcolor} 

\usepackage{fvextra} 
\DefineVerbatimEnvironment{myverbatim}{Verbatim}
 {breaklines=true, fontsize=\small}

\usepackage{newrevisor}
\newrevisor{manuel}{violet!75}
\newrevisor{guorui}{blue!75}
\newrevisor{mingfei}{red!75}
\hiderevisor{manuel}
\hiderevisor{guorui}

\title{Re-evaluating LLM-based Heuristic Search: A Case Study on the 3D Packing Problem}

\author{Guorui Quan}
\email{guorui.quan@postgrad.manchester.ac.uk}
\affiliation{%
  \department{Department of Computer Science}
  \institution{The University of Manchester}
  \city{Manchester}
  \country{United Kingdom}
}

\author{Mingfei Sun}
\email{mingfei.sun@manchester.ac.uk}
\affiliation{%
  \department{Department of Computer Science}
  \institution{The University of Manchester}
  \city{Manchester}
  \country{United Kingdom}
}

\author{Manuel L\'{o}pez-Ib\'{a}\~{n}ez}
\email{manuel.lopez-ibanez@manchester.ac.uk}
\affiliation{%
  \department{Alliance Manchester Business School}
  \institution{University of Manchester}
  \city{Manchester}
  \country{United Kingdom}
}
\begin{document}

\begin{abstract}
The art of heuristic design has traditionally been a human pursuit. While Large Language Models (LLMs) can generate code for search heuristics, their application has largely been confined to adjusting simple functions within human-crafted frameworks, leaving their capacity for broader innovation an open question.
To investigate this, we tasked an LLM with building a complete solver for the constrained 3D Packing Problem. Direct code generation quickly proved fragile, prompting us to introduce two supports: constraint scaffolding—prewritten constraint-checking code—and iterative self-correction—additional refinement cycles to repair bugs and produce a viable initial population. Notably, even within a vast search space in a greedy process, the LLM concentrated its efforts almost exclusively on refining the scoring function. This suggests that the emphasis on scoring functions in prior work may reflect not a principled strategy, but rather a natural limitation of LLM capabilities. The resulting heuristic was comparable to a human-designed greedy algorithm, and when its scoring function was integrated into a human-crafted metaheuristic, its performance rivaled established solvers, though its effectiveness waned as constraints tightened. Our findings highlight two major barriers to automated heuristic design with current LLMs: the engineering required to mitigate their fragility in complex reasoning tasks, and the influence of pretrained biases, which can prematurely narrow the search for novel solutions.

\end{abstract}

\maketitle

\section{Introduction}
The art of heuristic design, a cornerstone of solving computationally hard problems, has traditionally been a human pursuit, blending intuition with deep domain expertise.
\guorui{}{This has produced a rich library of effective strategies for packing problems, from classic sorting-based rules for one-dimensional bin packing~\citep{coffman2013bin} to complex spatial placement algorithms for 2D and 3D variants~\citep{ali2022line}. Automating this discovery process has long been a goal in the field; for decades, methods like Genetic Programming have successfully evolved heuristics by applying evolutionary principles to populations of programs that represent heuristics~\citep{ahvanooey2019survey}.} 
The recent emergence of methods like FunSearch~\citep{romera2024mathematical} and Evolution of Heuristics (EoH)\citep{liu2024evolution} \guorui{}{represents a new paradigm in this quest}, demonstrating that Large Language Models (LLMs) can discover novel, and sometimes superior, solutions to complex combinatorial problems. 

However, these landmark successes share a crucial context: the LLM is typically tasked with optimizing a small, well-defined component within a mature, human-crafted algorithmic framework. For instance, in tackling the Traveling Salesperson Problem (TSP), the LLM's role was confined to refining the scoring logic of a proven 2-Opt metaheuristic, not inventing a routing algorithm from first principles\citep{liu2024evolution}. This paradigm thrives in "knowledge-rich" environments where a robust optimization framework already exists, raising a critical, open question: can LLMs innovate more broadly when this expert framework is removed, forcing them to confront the full, unstructured complexity of a problem on their own?

To probe these limits, we tasked an LLM with a new challenge: building a complete solver from scratch for the constrained 3D Packing Problem~\citep{bortfeldt2013constraints}. Characterized by high-dimensional geometry and real-world constraints, this problem provides a ``knowledge-poor'' environment, ideal for testing an LLM's autonomous discovery capabilities. \guorui{}{Our first experiment shows that applying the EoH framework~\citep{liu2024evolution} to a minimal greedy solver, where the goal is to discover a heuristic function that constructively builds the solution}, fails to produce valid program consistently, halting the evolutionary process before it could even begin due to logical errors, invalid geometric assumptions, and unhandled edge cases.

\guorui{}{To overcome these issues,} we introduced two critical supports to bridge the gap between the LLM's generative capabilities and the problem's logical demands. First, we developed constraint scaffolding, a pre-written library of verified constraint-checking code that offloads the burden of complex, error-prone geometric reasoning from the LLM, allowing it to focus on high-level strategy. Second, we implemented an iterative self-correction loop, a process where the LLM is repeatedly shown its own runtime errors and given the chance to repair its code, ensuring the generation of a viable initial population of heuristics.

With these supports in place, the LLM was able to produce a set of functional packing heuristics. \guorui{}{By analyzing the generated heuristics, we observed that} despite the vast search space available within a greedy process, \guorui{}{the iterative refinements of the heuristic generated by the LLM are almost exclusively concentrated on a single component}: the scoring function. It did not invent novel algorithmic structures like wall-building~\citep{che2011multiple}. Instead, it learned to combine familiar concepts like volume, surface area, and stability into increasingly sophisticated mathematical expressions to guide a simple placement strategy. This finding suggests that the focus on scoring functions in prior work may reflect not a principled strategy, but rather a natural limitation of current LLMs, which appear biased toward refining modular, mathematical components over inventing complex procedural logic.

This LLM-discovered scoring function, born from a simple greedy process, proved to be comparable to a standard human-designed greedy algorithm. When we integrated it into a more sophisticated metaheuristic, its performance reached the level of established, state-of-the-art solvers on standard benchmark tests. However, as tighter constraints such as load stability and item separation were introduced, the heuristic’s effectiveness waned slightly, causing the performance gap to leading solvers to increase. This suggests its success was heavily reliant on the human-predefined metaheuristic, which had been specifically designed for the initial, less-constrained problem.

Our findings highlight two major barriers to fully automated heuristic design with current LLMs. The first is the significant engineering required to mitigate their inherent fragility in tasks demanding complex, multi-step logical reasoning. The second is the influence of pretrained biases, which can prematurely narrow the search for novel algorithmic solutions by defaulting to familiar patterns.

\section{Related work}

\paragraph{Classical Approaches to the Constrained 3D Packing Problem}
The scope of the 3D packing problem has evolved significantly, moving beyond simple volume maximization to integrate a rich array of real-world constraints. These are broadly categorized into physical limitations—such as fragility~\citep{alonso2019mathematical, ramos2018new}, vertical stability~\citep{kurpel2020exact}, weight distribution for vehicle balance~\citep{moon2014container, ramos2018new}, and item orientation restrictions~\citep{ramos2018new}—and logistical requirements, including multi-drop (LIFO) sequencing~\citep{olsson2020automating}, complete shipment of orders~\citep{gimenez2021logistic, sheng2017heuristic}, and item grouping or separation~\citep{sheng2017heuristic}. To address this complexity, traditional solution methodologies have diverged into two main approaches.  

One approach involves exact methods, typically employing Integer Linear Programming (ILP) on discretized grids. While these methods can find provably optimal solutions~\citep{kurpel2020exact}, their computational demands make them prohibitive for time-sensitive, large-scale operations~\citep{zhu2020integer}. Consequently, a substantial body of work focuses on expert-designed heuristics and metaheuristics. These methods offer a trade-off between solution quality and speed, ranging from constructive tree search algorithms~\citep{eley2002solving} and Greedy Randomized Adaptive Search Procedures (GRASP)~\citep{che2011multiple} to powerful two-stage frameworks based on column generation~\citep{che2011multiple, zhu2012prototype} or set partitioning~\citep{eley2003bottleneck}. While effective, these highly-tuned methods are often challenging to adapt to novel combinations of constraints, highlighting the need for more automated discovery paradigms.

\paragraph{Automated Heuristic Discovery with Large Language Models}
In recent years, significant attention has been directed toward leveraging the capabilities of Large Language Models (LLMs) to solve complex optimization problems. This effort has mainly followed three approaches. 
The first approach uses the LLM directly as an optimizer~\citep{yang2023large}. The LLM receives a prompt describing the problem to proposes new solutions, along with a history of past attempt solutions and their scores. These are evaluated and fed back into the prompt, creating an iterative loop that gradually refines the results. While effective on small problems, performance drops sharply on larger ones due to context window limits. The high computational cost of each step also hinder the model’s ability to track complex relationships at scale.
A second approach involves using the LLM to generate formal mathematical models, which can then be fed into off-the-shelf mathematical programming solver~\citep{ahmaditeshnizi2024optimus}. However, Achieving good results typically requires a complex, multi-agent iterative process involving multiple calls to the LLM, making the approach less practical for interactive or rapid-prototyping scenarios compared to a human expert.
Moreover, this method inherits the scalability issues inherent in the exact optimization methods themselves.

The third, and arguably most promising, approach integrates the LLM within an evolutionary search framework. Methods like FunSearch~\citep{romera2024mathematical} and Evolution of Heuristics (EoH)~\citep{liu2024evolution} represent a promising frontier in this domain. This paradigm is best understood not as generating entire algorithms from scratch, but as a ``component optimization'' model. Within this framework, the LLM functions as a highly sophisticated mutation operator embedded within a fixed, human-designed algorithmic structure. Its role is to evolve a small, creative piece of code---a ``component''---that seamlessly integrates into and enhances this larger, pre-existing framework. For instance, The EoH method tackles the Traveling Salesman Problem (TSP) by discovering a more effective cost function to guide a mature, expert-designed meta-heuristic like 2-Opt~\citep{liu2024evolution}. Similarly, extensions like ReEvo enhance existing algorithms by evolving components such as penalty heuristics for Guided Local Search (GLS) or heuristic measures within the Ant Colony Optimization (ACO) framework~\citep{ye2024reevo}.

This observation, however, highlights a crucial and recurring theme: the paradigm's success has been demonstrated almost exclusively on thoroughly-studied and relatively simple problems, where not only are effective algorithmic structures already known, but generating feasible solutions is not a significant challenge in itself. The core question of whether this LLM-driven "component optimization" approach can be successfully applied to more complex, real-world domains—where a powerful expert framework may not be readily available and feasibility itself is a major hurdle—remains largely unanswered.

\section{Preliminaries}
\label{sec:preliminaries}

In this section, we first present a formal definition of the Constrained 3D Packing Problem, detailing its core objective and the real-world complexities. We then describe the Evolution of Heuristics (EoH) framework, the LLM-based search algorithm that serves as the basis for our investigation.

\subsection{The Constrained 3D Packing Problem}

We address the Input Minimization variant of the 3D packing problem, based on the formulation by \citet{kurpel2020exact}. The central goal is to pack a given collection of items into the minimum possible number of containers while satisfying a set of geometric constraints.

\paragraph{Core Problem Definition.}
The inputs consist of a set of items to be packed $J$. These items are categorized into a set of box types $I$. Each box type $i \in I$ is defined by its dimensions $(l_i, w_i, h_i)$, and a set of permissible orthogonal orientations $\Omega_i$. Additionally, there is a set of available container types $K$, each with internal dimensions $(L_k, W_k, H_k)$. A solution is a packing plan that assigns each item $j \in J$ to a specific container, an orientation $g_j \in \Omega_j$, and a continuous coordinate $(p_j, q_j, r_j)$ for its back-bottom-left corner. A plan is considered valid if it satisfies two fundamental constraints: (1) all items are fully contained within their assigned container boundaries, and (2) no two items within the same container overlap. The primary objective is to find a valid solution that minimizes the total number of containers used.

\newcommand{\Support}{\textit{Support}}

\paragraph{Real-World Constraints.}
To model realistic industrial scenarios, we incorporate two additional complex constraints from \citet{kurpel2020exact}. First, we introduce an incompatibility constraint where, given two disjoint sets of box types $B_1, B_2 \subset I$ (which do not necessarily cover all of $I$), no single container may hold items with box types from both sets simultaneously. Second, we enforce vertical stability. For any item $j$ not resting on the container floor, its base must be sufficiently supported by the items directly beneath it. Formally, given a stability coefficient $\alpha \in (0, 1]$, the total overlap area with its supporting items, denoted by the set $\Support(j)$, must be at least $\alpha$ times its own base area: $\alpha\cdot l'_j\cdot w'_j \leq \sum_{j' \in \Support(j)} A(j, j')$,
 where $A(j,j')$ is the overlap area and $(l'_j, w'_j)$ are the base dimensions of box $j$ in its current orientation.

\subsection{The Evolution of Heuristics (EoH) Framework}

Our work leverages the LLM-driven evolutionary search paradigm to automate algorithm discovery. The system begins with a collection, or "population," of simple computer programs that attempt to solve a problem. In each cycle, these programs are tested and scored for performance. The best ones are then shown to an LLM with a prompt asking it to create a new, improved version. The LLM acts as a sophisticated "mutation operator," writing a new program that is then added back to the population. This iterative loop of evaluation, selection, and LLM-driven generation allows the system to progressively discover highly effective heuristics. For a comprehensive technical description, we refer readers to the original work~\citep{romera2024mathematical}.

Building on this foundation, the Evolution of Heuristics (EoH) framework~\citep{liu2024evolution} introduces a refinement: the dual representation of each candidate heuristic as both a natural language description (a "thought") and its corresponding executable code. This allows the LLM to reason not just about syntax, but also about the underlying strategic intent. The evolutionary cycle selects parent (thought, code) pairs, uses them to prompt the LLM to generate a new child heuristic, and manages the population based on fitness scores.
To drive this process, EoH employs a suite of five prompt strategies that function as specialized genetic operators:

\paragraph{Exploration Operators (E1, E2):} These prompts are designed to introduce novel ideas into the population.
\begin{itemize}
    \item \textbf{E1 (Diversify):} Prompts the LLM to create a new heuristic that is as different as possible from the provided parents.
    \item \textbf{E2 (Synthesize):} Asks the LLM to identify a common theme among parents and then generate a new heuristic that builds upon this shared idea.
\end{itemize}

\paragraph{Modification Operators (M1, M2, M3):} These prompts focus on incrementally refining existing heuristics.
\begin{itemize}
    \item \textbf{M1 (Improve):} Instructs the LLM to make general performance enhancements to a single parent heuristic.
    \item \textbf{M2 (Tune):} Focuses the LLM on adjusting specific parameters within a heuristic's code.
    \item \textbf{M3 (Simplify):} Directs the LLM to remove redundant or unnecessary components to create a more elegant solution.
\end{itemize}

At each generation, each strategy is called to generate new heuristics. Each newly generated heuristic will be evaluated on problem instances and added to the current population if it is feasible.

\section{Enabling Heuristic Search for Complex Problems}

In this section, we first outline the baseline approach of applying EoH directly. We then articulate the fundamental challenges this approach faces—namely, the high frequency of logical, geometric, and efficiency errors in the LLM-generated code. Motivated by these challenges, we introduce two interventions designed to make the LLM-driven search process viable and effective: \emph{Constraint Scaffolding} and \emph{Iterative Self-Correction}.

\subsection{EoH for 3D packing problem}

As a first step, we apply the EoH framework directly to the 3D packing problem. Following the methodology of the 1D Bin Packing problem in the original EoH paper, we task an LLM with generating a single, monolithic Python function, \texttt{place\_item}. This function, embedded within a greedy constructive framework, is solely responsible for all decision-making: selecting an item, a container, a position, and an orientation, based on the items already placed and those not yet placed. The complete prompt provided to the LLM for this task is detailed in Appendix~\ref{app:naive_prompt}. As our experiments will show, this direct approach highlights significant challenges in applying EoH to domains with intricate physical and logical constraints.

The complexity that arises from switching from a 1D packing problem to a 3D packing problem places an immense burden on the LLM. It struggles to reliably generate code that is correct, valid, and efficient. The generated programs are frequently plagued by three categories of issues:
\begin{enumerate}
    \item \textbf{Logical and Runtime Errors:} We found once we require LLMs to generate a long code with complex logic, the programming mistakes will increase, such as forgetting to define return variables for every logic path.
    \item \textbf{Constraint Violations:} Failures in writing code that handles the constraints correctly, leading to geometrically impossible solutions (e.g., item collisions or out-of-bounds placements).
    \item \textbf{Computational Inefficiency:} The generation of naive algorithms (e.g., brute-force search) that fail to complete within a reasonable time limit.
\end{enumerate}

These issues, which are less prevalent or absent in problems with simpler formulations and fewer constraints, become critical in a complex domain like 3D packing. The original EoH methodology, when faced with such invalid or inefficient programs, typically discards them. This approach was viable for simpler problems where these issues were rare. However, in more complex problems, the dramatic increase in these types of errors significantly impedes the evolutionary process. It makes population mutation difficult and can even prevent the formation of a sufficient initial population of valid programs.

\subsection{Intervention 1: Constraint Scaffolding}

Our primary challenge was to mitigate the logical and geometric errors common in LLM-generated code. We observed that the frequency of these errors increases with code length and complexity. Since the LLM is required to implement a complete solver, it must interleave high-level strategic logic with low-level, deterministic routines for constraint validation (e.g., overlap detection, stability analysis). While essential for a valid solution, implementing these routines is ancillary to the core heuristic design and introduces frequent opportunities for error.

Our first intervention, Constraint Scaffolding, addresses this by enforcing a principled separation of concerns. Instead of tasking the LLM with generating a full implementation from scratch, we provide it with a verified API that encapsulates all necessary geometric and physical constraint checks. This elevates the LLM's role from a low-level coder to a high-level strategist, responsible only for orchestrating calls to this API to define a novel heuristic. This approach not only curtails a primary source of logical and geometric errors but also focuses the LLM’s generative capabilities on the creative aspects of problem-solving, effectively separating heuristic innovation from the deterministic mechanics of validation. In practice, we implement this API as a Python base class, \texttt{BaseAlgorithm}, which encapsulates the complex logic for overlap detection, stability checks, and item separation. The LLM is then tasked with implementing a new class that inherits from \texttt{BaseAlgorithm}, using its methods to orchestrate a high-level packing strategy.

\subsection{Intervention 2: Iterative Self-Correction}

While our constraint scaffolding corrects logical and semantic errors, it cannot directly address execution timeouts, which arise when a generated heuristic is too computationally expensive for its allocated time budget. In frameworks like the Evolution of Heuristics (EoH), such computationally expensive solutions are typically discarded. This approach acts as an implicit penalty for inefficiency but offers no direct feedback to the generative process. This passive penalization is insufficient for an LLM-based generator. Unlike stochastic search, an LLM may exhibit systematic biases towards certain algorithmic structures; without explicit feedback on performance, it may repeatedly generate similarly inefficient solutions rather than exploring more performant alternatives.

To overcome this, we introduce an iterative self-correction mechanism, analogous to the "repair mechanisms" commonly used in genetic programming~\citep{michalewicz1996heuristic}. When a candidate program fails—due to (1) a Python SyntaxError, (2) a constraint violation, or (3) an execution timeout—our system captures the relevant diagnostic output. This feedback, such as the full traceback, a structured constraint violation report, or a timeout notification, is appended to the prompt. The LLM is then instructed to rectify the fault. Specifically, it is prompted to implement a new Python class that inherits from the faulty one, overriding only the methods necessary to correct the error. This correction cycle can be repeated for up to five attempts, allowing the model to progressively refine its code into a viable solution.
This process serves a dual purpose: it directly enables the generation of more computationally efficient heuristics by addressing timeouts, and it increases the overall robustness of the population generation by reducing the number of invalid candidates.

\section{Experiments}
\label{sec:experiments}

\subsection{Experimental Setup}

\paragraph{Tasks and Datasets.}
We evaluate our method on the 47 Single Stock-Size Cutting Stock Problem (SSSCSP) instances from the benchmark dataset introduced by \cite{kurpel2020exact}. As described in the problem definition, the objective is to minimize the number of containers used. We select the first 20 instances as our training set, where the evolutionary search is performed to discover a heuristic. The remaining 27 instances serve as the unseen test set to evaluate the generalization of the discovered heuristic. We test performance on two distinct variants of the problem:
\begin{itemize}
    \item \textbf{Base Problem:} The standard 3D packing problem without the additional practical constraints.
    \item \textbf{Constrained Problem:} The 3D packing problem with the full set of constraints, including stability and the separation of incompatible boxes. For the stability constraint, we set the coefficient $\alpha = 1.0$, requiring full base support for all boxes. This setting is chosen to directly align with the experiments in \cite{kurpel2020exact}, ensuring a fair comparison.
\end{itemize}

\paragraph{Implementation Details.}
Our method is built on the EoH~\citep{liu2024evolution} paradigm. We use Gemini-2.0-flash as the code-generation model. The evolutionary process is run for 10 rounds on the training set. All experiments were conducted on a server equipped with an AMD EPYC 7B12 CPU (2.2 GHz, up to 3.3 GHz turbo) and 128 GB of RAM. The discovered packing heuristics are implemented in single-threaded Python.

\paragraph{Baselines.} To assess performance, we compare against a comprehensive set of human-designed methods from the literature, including greedy heuristics (\textbf{First-Fit}, \textbf{S-GRASP}), advanced metaheuristics (\textbf{IVA}, \textbf{ZHU}), and a state-of-the-art \textbf{Exact Solver} that provides the optimal solution. A full description of these methods is in Appendix~\ref{app:benchmark_details}.

\subsection{Validating the Interventions}

Before presenting our main results, we first validate the necessity of our two primary interventions: constraint scaffolding and iterative self-correction. We compare our full framework against a basic EoH baseline, which represents a direct application of the EoH paradigm without our proposed enhancements.

To isolate the effect of the interventions, we generated an initial population of 20 candidate programs using the basic EoH baseline and evaluated each on a training set. The result was a near-total failure: 90\% of the generated programs were unable to produce solution in all instances in the training dataset. A manual analysis of the failures identified three primary categories:

\begin{itemize}
\item \textbf{Runtime Errors (45\%):} The most common failure, occurring when the LLM generated complex conditional logic but failed to initialize key variables on all code paths—a fundamental programming mistake.
\item \textbf{Invalid Solutions (35\%):} Cases where the code executed but produced geometrically impossible placements (e.g., collisions) or returned incorrect values for some decision variables (e.g., \texttt{None} for an orientation), highlighting a failure to implement spatial reasoning.
\item \textbf{Timeouts (10\%):} Programs that failed to complete, consistent with the LLM attempting computationally infeasible algorithms such as a grid search.
\end{itemize}

Next, we evaluated the cumulative impact of our interventions. We generated another initial population of 20 candidates, this time using our constraint scaffolding. We then applied our iterative self-correction process for three rounds. Table~\ref{tab:self_correction_effectiveness} tracks the number of feasible heuristics at each stage. We group "Runtime Errors" and "Infeasible Solutions" from the previous analysis under the broader category of "Code Error."

\begin{table}[h!]
\centering
\caption{Effectiveness of Scaffolding and Iterative Self-Correction vs.\ Naive Baseline. Cumulative count of valid vs.\ failed heuristics for an initial population of 20 candidates.}
\label{tab:self_correction_effectiveness}
\begin{tabular}{lccc}
\toprule
\textbf{Approach} & \textbf{Valid} & \textbf{Timeout} & \textbf{Code Error} \\
\midrule
Baseline & 2 & 2 & 16 \\
\midrule
Scaffolded (Initial Gen.) & 6 & 10 & 4 \\
\quad + Iteration 1        & 8 & 10 & 2 \\
\quad + Iteration 2        & 15 & 4 & 1 \\
\quad + Iteration 3        & 17 & 3 & 0 \\
\bottomrule
\end{tabular}
\end{table}

The results show that both interventions are critical:
\begin{enumerate}
    \item \textbf{Constraint Scaffolding} provides an immediate and dramatic benefit, reducing code errors from $16 \rightarrow 4$. It ensures \emph{logical correctness} by abstracting away low-level geometric calculations, but it does not guarantee \emph{computational efficiency}, as evidenced by the high number of timeouts (10).
    
    \item \textbf{Iterative Self-Correction (ISC)} then systematically resolves the remaining issues. It eliminates all remaining code errors and significantly reduces timeouts, increasing the feasible count from $6 \rightarrow 17$ (an 85\% success rate). While we did not test ISC on the naive baseline directly, these results strongly suggest that scaffolding is a \emph{necessary prerequisite} for ISC to be effective, as the self-correction process is better suited to fixing high-level algorithmic inefficiencies than fundamental geometric errors.
\end{enumerate}
In short, scaffolding makes the problem \emph{solvable}, while self-correction makes the solutions \emph{efficient}.

\subsection{Heuristic Discovery on the Base Problem}

\begin{figure}[htbp]
\centering
\includegraphics[width=0.6\textwidth]{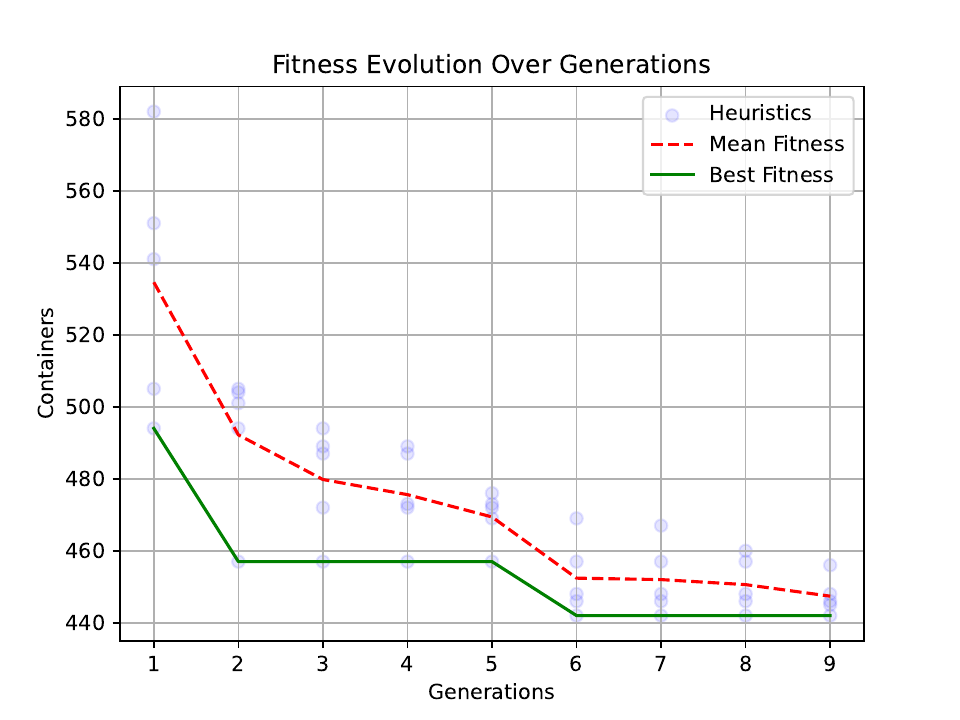}
\caption{The evolutionary trajectory for one run, plotting the fitness of the best-performing heuristic in each generation evaluated on training datasets.}
\label{fig:evolution}
\end{figure}

\paragraph{Evolutionary Trajectory and Performance.}
We now applied our full framework (EoH with Scaffolding and ISC) to discover a packing heuristic for the base problem. We ran the evolutionary process using two different time budgets for heuristic evaluation: 10 and 60 seconds. This budget represents the total time allowed for a single candidate heuristic to solve all 20 instances in the training set. With a population size of 5 and 10 generations, a full evolutionary run using the 10-second budget is computationally inexpensive, taking approximately 30 minutes, excluding the overhead for LLM calls. Figure~\ref{fig:evolution} shows a typical evolutionary trajectory, plotting the fitness (average containers used on the training set) of the best-performing heuristic in each generation. The search effectively improves the population's fitness, converging towards a high-quality solution around the sixth generation.

As shown in Table~\ref{tab:base_problem_results}, the best heuristic discovered by our EoH process outperforms the naive First-Fit baseline and is competitive with the S-GRASP heuristic. However, it still trails state-of-the-art methods like ZHU by a significant margin. Further analysis points to two limitations. First, the LLM-discovered heuristic shows signs of overfitting, as its performance gap relative to the exact solver widens from the training to the test set. Second, the search appears to be limited by the heuristic space rather than computation time, as increasing the budget from 10s to 60s yielded no significant improvement. A detailed, instance-by-instance comparison is available in Appendix~\ref{app:detailed_results}.

\begin{table}[h!]
\centering
\caption{Performance on the Base Problem. The "Total" column is the sum of containers used across the training and test sets. Lower is better. EoH results are averaged over three random seeds, with the single best-performing discovered heuristic shown separately.}
\label{tab:base_problem_results}
\begin{tabular}{lcccc}
\toprule
\textbf{Method}                      & \textbf{Budget} & \textbf{Train} & \textbf{Test} & \textbf{Total} \\ 
\midrule
\multicolumn{5}{c}{\textit{\textbf{Human-Designed Heuristics}}} \\
\midrule
\quad IVA \citep{ivancic1988integer} & - & 446 & 317 & 763 \\
\quad BOR \citep{bortfeldt2000heuristik} & - & 416 & 289 & 705 \\
\quad ELY \citep{eley2003bottleneck} & - & 412 & 287 & 699 \\
\quad LZG \citep{lim2005container} & - & 410 & 284 & 694 \\
\quad GRMod \citep{moura2005grasp}   & -                   & 410            & 283           & 693            \\ 
\quad CHE \citep{che2011multiple} & - & 410 & 282 & 692 \\
\quad ZHU \citep{zhu2012prototype} & - & 410 & 281 & 691 \\
\quad Exact Solver \citep{kurpel2020exact} & - & \textbf{409} & \textbf{280} & \textbf{689} \\
\quad First-Fit(Greedy)      & -                   & 495            & 329           & 824            \\
\quad S-GRASP(Greedy) \citep{che2011multiple}       & -                   & 437            & 325           & 762            \\
\midrule
\multicolumn{5}{c}{\textit{\textbf{LLM-Discovered (EoH)}}} \\
\midrule
\quad Avg. Performance               & 10s                 & $443.7\pm3.8$            & $333.7\pm1.5$           &  $777.3\pm4.2$            \\
\quad Avg. Performance               & 60s                 &   $444.3\pm5.9$          & $336.7\pm15.2$           & $781.0\pm21$            \\
\quad Best Discovered Heuristic & 60s        & 440   & 323  & 763   \\ 
\bottomrule
\end{tabular}
\end{table}

\paragraph{Analysis: The LLM as a Scoring Function Optimizer.}

To understand \emph{why} the LLM-discovered heuristics perform as they do, we analyzed the code generated throughout the evolutionary process. Our analysis reveals that the LLM does not discover fundamentally new algorithmic structures but rather refines a narrow set of familiar strategies by evolving increasingly complex scoring functions.

The process began with a homogeneous population of simple "largest-item-first" heuristics—a basic greedy approach the LLM appears biased toward. From this starting point, the LLM never invented novel high-level concepts like wall-building~\citep{che2011multiple}. Rather, its innovations were concentrated in two key areas: the criteria for selecting the next item and the logic for evaluating its placement. Initially, simple rules like item volume governed selection. This evolved to incorporate more factors, such as an box's quantity, its ``cubeness'', and how well it might fit into a nearly-full container. Simultaneously, the placement logic transformed from a simple \emph{first-fit} to a \emph{best-fit} strategy. The LLM developed complex scoring functions that learned to minimize wasted space, build level layers by reducing overall height, and maximize surface contact for denser packing. For specific examples of these evolved functions, see Appendix~\ref{app:scoring_functions}.

\subsection{Evaluating the Discovered Scoring Function}

The analysis above reveals a crucial insight: the LLM excels at optimizing scoring functions, but the performance of its discovered heuristic is ultimately bottlenecked by the simple greedy framework in which it operates. This finding prompted a follow-up investigation: is the discovered scoring function genuinely powerful, with its potential being artificially capped by its simple environment?

To test this, we transplanted the LLM's best scoring function into a sophisticated, human-designed search architecture. We hypothesized that freed from its greedy constraints, the LLM's heuristic would compete with state-of-the-art approaches. This would prove that the LLM is a powerful component optimizer, and that the success of many LLM-driven methods may rely heavily on the advanced search frameworks they are built upon.

To test our hypothesis, we developed a hybrid method named \textbf{LLM-Score + RS + SP}. This method extracts the best-performing scoring function from our earlier experiments and ``transplants'' it into a classic two-stage metaheuristic framework from Operations Research~\citep{eley2003bottleneck}:
\begin{enumerate}
    \item \textbf{Randomized Search (RS):} The LLM-generated scoring function guides a randomized search to generate a large and diverse pool of high-quality candidate packing patterns.
    \item \textbf{Set Partitioning (SP):} An exact SP model selects the optimal subset of these patterns to fulfill all orders using the minimum number of containers.
\end{enumerate}

To introduce diversity in the search, the RS stage employs a randomization parameter, $\beta$. For each item placement decision, instead of greedily choosing the item with the maximum score, we construct a candidate list of all items whose scores fall within the range $[s_{\max} - \beta (s_{\max} - s_{\min}), s_{\max}]$, where $s_{\min}$ and $s_{\max}$ are the minimum and maximum scores of all items available for placement at this step. An item is then selected uniformly at random from this list.

\paragraph{Baselines and Implementation Details}

We benchmarked our hybrid method against two strong baselines that share the same RS + SP architecture but differ in their scoring component:
\begin{itemize}
    \item \textbf{S-GRASP + RS + SP:} A human-designed baseline where the Randomized Search is guided by the S-GRASP heuristic (which has its own randomization parameter, $\delta$). This provides a direct comparison between the LLM-discovered component and a strong, human-designed counterpart.
    \item \textbf{Random + RS + SP:}  A control baseline where the scoring function is replaced by a constant value, making the selection purely random. This helps quantify the contribution of the LLM's intelligent guidance over an unguided search.
\end{itemize}

We generated three distinct candidate pools for the set partition model. The Random baseline pool consisted of 10,000 runs, while the S-GRASP pool was created from 5,001 runs (500 for each $\delta \in \{0.1, \dots, 1.0\}$, plus one deterministic run). In contrast, our LLM-Score utilized a pool of just 401 runs, comprising one deterministic run ($\alpha=0.0$) and 200 randomized runs for each of $\alpha=0.05$ and $\alpha=0.15$. For robustness, all experiments were conducted with three different random seeds, and results are reported as mean $\pm$ standard deviation.\

\paragraph{Success on the Base Problem.}

As shown in Table~\ref{tab:hybrid_results}, our hybrid method achieved impressive results on the base problem. The \textbf{LLM-Score + RS + SP} method significantly outperformed both the original greedy heuristic and the S-GRASP-guided hybrid. It achieved near-optimal performance, nearly matching the state-of-the-art ZHU heuristic and the Exact Solver. A detailed, instance-by-instance comparison is provided in Appendix~\ref{app:detailed_results}. Critically, it accomplished this using a candidate pool of just 401 random runs, demonstrating high search efficiency. This result confirms our hypothesis: when placed within a powerful search framework, the LLM-generated scoring function proves to be a high-quality component, competitive with human-designed heuristics on the base problem.

\begin{table}[h!]
\centering
\caption{Performance on the Base Problem, comparing the LLM-discovered component within different frameworks.  Results are averaged over three random seeds. Lower is better.}
\label{tab:hybrid_results}
\begin{tabular}{lccc}
\toprule
\textbf{Method} & \textbf{Train} & \textbf{Test} & \textbf{Total} \\ 
\midrule
\multicolumn{4}{l}{\textit{\textbf{LLM-Discovered Heuristic}}} \\
\quad Best Heuristic from EoH & 440 & 323 & 763 \\
\midrule
\multicolumn{4}{l}{\textit{\textbf{Hybrid Methods (Score + RS + SP Framework)}}} \\
\quad \textbf{LLM-Score} & \textbf{410.0 $\pm$ 0.0} & \textbf{284.3 $\pm$ 0.22} & \textbf{694.3 $\pm$ 0.22} \\
\quad S-GRASP \citep{che2011multiple} & 413.0 $\pm$ 0.0 & 286.7 $\pm$ 0.22 & 699.7 $\pm$ 0.22 \\
\quad Random (Unguided Search) & 414.0 $\pm$ 0.0 & 300.0 $\pm$ 0.66 & 714.0 $\pm$ 0.66 \\
\midrule
\multicolumn{4}{l}{\textit{\textbf{Benchmarks}}} \\
\quad ZHU \citep{zhu2012prototype}  & 410 & 281 & 691 \\
\quad Exact Solver \citep{kurpel2020exact} & 409 & 280 & \textbf{689} \\
\bottomrule
\end{tabular}
\end{table}

\subsection{Results on the Constrained Problem}
Having established a baseline, we tested the robustness of the LLM-generated solutions on more difficult, constrained versions of the problem. This stage was critical to evaluating whether the LLM's narrow focus on scoring-function optimization could produce a generally effective heuristic. Table~\ref{tab:constrained_results_merged} presents the performance when adding load stability, item separation, or both constraints.  A detailed, instance-by-instance comparison is available in Appendix~\ref{app:detailed_results_constrained}. We found that the methods developed by the LLM all use a scoring system (as described in the previous section) and apply constraints by filtering out options that do not satisfy them.

The results reveal a weakness. While human-designed heuristics like ZST maintain near-optimal performance, the LLM-discovered heuristic and the hybrid method both falter significantly as constraints are added. The LLM's optimization efforts, which were narrowly focused on the scoring function, produced a heuristic that is brittle. While effective for the unconstrained problem, the scoring function alone is insufficient to navigate the complex trade-offs introduced by \manuel{hard}{real-world} constraints.

\begin{table}[h!]
\centering
\caption{Consolidated performance on constrained problems. "Total" is the sum of containers used across train and test sets (lower is better).}
\label{tab:constrained_results_merged}
\begin{tabular}{l l ccc}
\toprule
\textbf{Constraint Set} & \textbf{Method} & \textbf{Train} & \textbf{Test} & \textbf{Total} \\ 
\midrule
\multicolumn{5}{l}{\textit{\textbf{Load Stability}}} \\
& ELS \citep{eley2002solving} & 427 & 306 & 733 \\
& ELP \citep{eley2002solving} & 421 & 300 & 721 \\
& ZST \citep{zhu2012prototype} & 411 & 282 & 693 \\
& Exact Solver \citep{kurpel2020exact} & \textbf{410} & \textbf{283} & \textbf{693} \\
& Avg. LLM Discovered & $440.3\pm2.3$ & $334.3\pm1.2$ & $774.6\pm3.1$ \\
& Best LLM Discovered & 439 & 333 & 772 \\
& Avg. Hybrid & $410.6\pm0.2$ & $290.0\pm0.6$ & $700.7\pm0.9$ \\
& Best Hybrid & 411 & 289 & 700 \\
\midrule
\multicolumn{5}{l}{\textit{\textbf{Separation}}} \\
& ELY \citep{eley2003bottleneck} & 425 & 300 & 725 \\
& Exact Solver \citep{kurpel2020exact} & \textbf{423} & \textbf{292} & \textbf{715} \\
& Avg. LLM Discovered & $460.3\pm6.0$ & $346.7\pm6.1$ & $807.0\pm1.7$ \\
& Best LLM Discovered & 454 & 352 & 806 \\
& Avg. Hybrid & $431.0\pm0.0$ & $307.0\pm0.7$ & $738.0\pm0.7$ \\
& Best Hybrid & 431 & 306 & 737 \\
\midrule
\multicolumn{5}{l}{\textit{\textbf{Both Constraints}}} \\
& Exact Solver \citep{kurpel2020exact} & \textbf{424} & \textbf{296} & \textbf{720} \\
& Avg. LLM Discovered & $458.7\pm38.7$ & $372.3\pm7.2$ & $831.0\pm45.9$ \\
& Best LLM Discovered & 455 & 351 & 806 \\
& Avg. Hybrid & $435.0\pm0.0$ & $311.0\pm0.0$ & $746.0\pm0.0$ \\
& Best Hybrid & 435 & 311 & 746 \\
\bottomrule
\end{tabular}
\end{table}

\section{Limitations}

Our methodology, while effective, is constrained by two primary limitations that also define clear avenues for future research.

First, the approach demands significant engineering. The failure of our initial, naive approach underscores that current LLMs cannot autonomously handle the logical and geometric complexities of a problem like 3D packing. The necessity of our interventions—constraint scaffolding and iterative self-correction—highlights this fragility. To address this, future work could proceed in two directions. One path involves improving the model's intrinsic capabilities through methods like fine-tuning~\citep{ouyang2022training}, which could equip the LLM to handle such constraints more autonomously. An alternative path is to reframe the generation task itself, shifting from producing executable code to generating abstract representations, such as mathematical formulas for a dedicated solver, thereby leveraging the LLM's strengths while offloading precise logic to other tools~\citep{zhang2025or}.

Second, the generalizability of our findings is limited by the scope of our experiments. We focused on a single problem, the 3D packing problem, using a specific model and framework (Gemini-2.0-flash with EoH). It is therefore uncertain whether our core observations, such as the critical need for scaffolding and the LLM's preference for certain heuristic structures, apply to other complex optimization problems or more powerful language models. Broadening this investigation faces considerable practical hurdles. The most advanced models are often behind restrictive and expensive APIs, limiting accessibility for academic research. Furthermore, the inherent randomness of LLM outputs and the high sensitivity to prompt engineering make it a non-trivial challenge to establish a fair and controlled basis for comparison across different models and methods\citep{liang2022holistic,gan2023sensitivity}.

\section{Conclusion}
Our exploration into the LLM's capacity for autonomous heuristic design in the 3D Packing Problem has yielded a set of findings. We first established that a naive application is infeasible, which led us to develop two significant engineering interventions—constraint scaffolding and iterative self-correction—to overcome the model's inherent fragility in complex geometric and logical reasoning. With this crucial support, the LLM-driven search successfully discovered a novel heuristic by focusing its optimization power on the scoring function of a simple greedy algorithm.

The practical consequences of this were compelling: the discovered scoring function was integrated into a human-designed metaheuristic to achieve strong, competitive performance on unconstrained benchmarks. While its effectiveness diminished when faced with realistic industrial constraints. Our findings identify two primary frontiers for automated heuristic design: first, advancing the engineering required to manage LLM fragility and, second, moving beyond the pretrained biases that currently steer the search toward familiar, component-level optimizations instead of truly novel algorithmic paradigms.

\bibliographystyle{plainnat}
\bibliography{sample}

\appendix

\section{Reproducibility Checklist}
\label{sec:appendix-checklist-simple}

\begin{itemize}
    \item \textbf{Code:} The complete source code is available at \url{https://github.com/quangr/EoH}.\MANUEL{This link is not working?}
    
    \item \textbf{Data:} We use the public SSSCSP dataset, which can be downloaded from \url{https://www.leandro-coelho.com/container-loading-problems/}. 
    \begin{itemize}
        \item The specific split into 20 training and 27 test instances is detailed in the repository's code.
        \item \textbf{Constraint Specification:} For the separation constraint experiments, items of the first and second type in each instance are defined as incompatible. This rule was established based on direct communication with the authors of \cite{kurpel2020exact} and is not specified in the original publication.
    \end{itemize}

    \item \textbf{Environment and Dependencies:}
    \begin{itemize}
        \item The code is written in Python 3.9.
        \item The Set Partitioning model is implemented using the PuLP library in Python.
    \end{itemize}
    
    \item \textbf{Hardware:} All experiments were conducted on a server with an AMD EPYC 7B12 CPU. No GPU is required.

    \item \textbf{LLM and Hyperparameters:}
    \begin{itemize}
        \item \textbf{Model:} Gemini-2.0-flash accessed via API.
        \item \textbf{Evolutionary Parameters:} The search was run for 10 rounds (generations) with a population size of 5\MANUEL{but for the initial experiments, you used a population size of 20, so why use 5 later? Does this choice limit the results? It seems quite small\ldots}. Specific prompts and parameters for mutation and crossover are provided in the appendix and the code.
        \item \textbf{Self-Correction:} The correction loop was run for a maximum of 5 attempts per candidate.
    \end{itemize}

\end{itemize}

\section{Prompts for Heuristic Generation}
\label{sec:appendix_prompts}

This appendix contains the exact, unedited prompts provided to the Large Language Model for the experiments.

\subsection{Naive Prompt for Standalone Function}
\label{app:naive_prompt}
The following prompt was used to generate standalone Python functions. It contains the problem definition, I/O specifications, and constraints, but provides no helper code or architectural guidance. This represents the "naive approach" baseline.

\begin{myverbatim}
You are to solve the **Single-Stock-Size Cutting Stock Problem (SSSCSP)**. You are given a list of item **types** (each with dimensions and a **quantity**) and the dimensions of a single standard container. Your goal is to pack all items into the minimum number of these identical containers.

Your task is to implement a single, standalone Python function named `place_item`. This function will be called repeatedly. In each call, it must analyze the remaining unplaced item types and the current state of all containers, and then decide on the single best placement to make next. This involves selecting an item type, a container for it, a position (x, y, z), and an orientation.

**Primary Objective: Minimize the total number of stock containers used.**

**Constraints:**
1.  **Single Container Type:** All containers are identical. No weight or support constraints apply.
2.  **Complete Placement:** All items of all types must be packed.
3.  **Item Orientation:** Items can be rotated. There are 6 possible orientations.
4.  **No Overlap:** Items cannot overlap.
5.  **Boundaries:** Items must be placed fully inside the container.
6.  **Immutable Inputs:** Your function must not modify its input arguments (`unplaced_items`, `trucks_in_use`). Treat them as read-only.

Data Distribution Summary:

Container Properties (identical for all containers in a problem):
  - Length: Min=10.00, 25th=16.25, Median=20.00, 75th=36.00, Max=60.00
  - Width: Min=6.00, 25th=12.50, Median=15.00, 75th=19.00, Max=40.00
  - Height: Min=14.00, 25th=16.00, Median=20.00, 75th=23.75, Max=72.00

Item Properties (based on all items in loaded instances):
  - Length: Min=2, 25th=8, Median=8, 75th=10, Max=40
  - Width: Min=4, 25th=6, Median=12, 75th=16, Max=32
  - Height: Min=4, 25th=5, Median=10, 75th=15, Max=24
First, describe your new algorithm and main steps in one sentence. The description must be inside a brace. Next, implement it in Python as a function named place_item. This function should accept 3 input(s): 'unplaced_items', 'trucks_in_use', 'truck_type'. The function should return 6 output(s): 'truck_index', 'item_index', 'x', 'y', 'z', 'orientation'. 
        **Input Arguments:**
        `unplaced_items`: A list of dictionaries, each representing an **item type**.
            - 'item_id' (str): The unique ID for this item type.
            - 'length', 'width', 'height' (float): The base dimensions of this item type.
            - 'quantity' (int): The number of items of this type still needing to be packed.

        `trucks_in_use`: A list of dictionaries, where each dictionary represents a container currently in use.
            - `'occupied_volumes'`: A list of **dictionaries**. Each dictionary represents a placed item and has the following keys:
                - `'item_type_id'` (str): The unique identifier for the *type* of the placed item.
                - `'x'` (float): The starting position along the container's length (x-axis).
                - `'y'` (float): The starting position along the container's width (y-axis).
                - `'z'` (float): The starting position along the container's height (z-axis).
                - `'length'` (float): The dimension of the item along the x-axis **in its placed orientation**.
                - `'width'` (float): The dimension of the item along the y-axis **in its placed orientation**.
                - `'height'` (float): The dimension of the item along the z-axis **in its placed orientation**.

        `truck_type`: A **single tuple** for the standard container dimensions: `(length, width, height)`.
            
        **Return Values:**
        Your function must return a tuple of 6 values: (`truck_index`, `item_index`, `x`, `y`, `z`, `orientation`)
        
        `truck_index` (int): Index of the container in `trucks_in_use`. Use **-1** to open a new container.
        `item_index` (int): Index of the item **type** in the `unplaced_items` list that you have decided to place.
        `x`, `y`, `z` (float): Position for the new item. If `truck_index` is -1, this is the position within the new container.
        `orientation` (int): An integer from 0 to 5, specifying the item's orientation. Let the original dimensions from `unplaced_items` be (L, W, H). The placed dimensions (length, width, height) will be:
            - 0: (L, W, H)
            - 1: (L, H, W)
            - 2: (W, L, H)
            - 3: (W, H, L)
            - 4: (H, L, W)
            - 5: (H, W, L)
         Your function must be self-contained and deterministic. A good strategy is crucial for a high score. Do not include any comments or print statements in your function.
Do not give additional explanations.
\end{myverbatim}

\subsection{Scaffolded Prompt for Class-Based Heuristic}
\label{app:scaffolded_prompt}
To guide the LLM towards more structured and logically sound solutions, the following prompt was used. It introduces a `BaseAlgorithm` class containing pre-verified, low-level geometric functions and instructs the LLM to implement a child class. This represents our first intervention, "Constraint Scaffolding".

\begin{myverbatim}
You are to solve the **Single-Stock-Size Cutting Stock Problem (SSSCSP)**. You are given a list of item **types** (each with dimensions and a **quantity**) and the dimensions of a single standard container. Your goal is to pack all items into the minimum number of these identical containers.
Design a novel algorithm that, at each step, selects an available item type, **chooses one of 6 possible orientations**, and finds a valid position for it in a container.
        **Primary Objective: Minimize the total number of stock containers used.**

        Constraints:
        1. **Single Container Type:** All containers are identical. No weight or support constraints apply.
        2. **Complete Placement:** All items of all types must be packed.
        3. **Item Orientation:** Items can be rotated. There are 6 possible orientations.
        4. **No Overlap:** Items cannot overlap.
        5. **Boundaries:** Items must be placed fully inside the container.
        6. **Immutable Inputs:** Your function must not modify its input arguments (unplaced_items, trucks_in_use). Treat them as read-only. The calling environment manages state.
        
Data Distribution Summary:

Container Properties (based on all loaded instances):
  - Length: Min=10.00, 25th=19.50, Median=22.50, 75th=35.00, Max=60.00
  - Width: Min=6.00, 25th=14.75, Median=17.00, 75th=26.25, Max=52.00
  - Height: Min=14.00, 25th=20.00, Median=24.00, 75th=32.50, Max=72.00

Item Properties (based on all items in loaded instances):
  - Length: Min=2.00, 25th=6.00, Median=8.00, 75th=14.25, Max=40.00
  - Width: Min=4.00, 25th=6.00, Median=12.00, 75th=15.75, Max=32.00
  - Height: Min=4.00, 25th=8.00, Median=10.00, 75th=14.25, Max=35.00
  - Quantity: Min=12.00, 25th=25.00, Median=32.50, 75th=45.00, Max=60.00

You will be implementing a class `Algorithm` that inherits from the following `BaseAlgorithm` class. You can use the helper methods provided in the base class.
```python
import numpy as np

class BaseAlgorithm:
    def __init__(self, epsilon=1e-6):
        # Initializes the base algorithm with common parameters.
        if not epsilon > 0:
            raise ValueError("epsilon should be a small positive value.")
        self.epsilon = epsilon

    def _check_overlap_3d(self, item1_pos, item1_dims, item2_pos, item2_dims):
        x1, y1, z1 = item1_pos
        l1, w1, h1 = item1_dims
        x2, y2, z2 = item2_pos
        l2, w2, h2 = item2_dims
        return (x1 < x2 + l2 and x1 + l1 > x2 and
                y1 < y2 + w2 and y1 + w1 > y2 and
                z1 < z2 + h2 and z1 + h1 > z2)

    def _get_orientations(self, item_type):
        L, W, H = item_type['length'], item_type['width'], item_type['height']
        return [
            (0, (L, W, H)), (1, (L, H, W)), (2, (W, L, H)),
            (3, (W, H, L)), (4, (H, L, W)), (5, (H, W, L))
        ]

    def _is_within_container_bounds(self, item_pos, item_dims, container_dims):
        px, py, pz = item_pos
        pl, pw, ph = item_dims
        cl, cw, ch = container_dims
        return (px >= 0.0 - self.epsilon and px + pl <= cl + self.epsilon and
                py >= 0.0 - self.epsilon and py + pw <= cw + self.epsilon and
                pz >= 0.0 - self.epsilon and pz + ph <= ch + self.epsilon)

    def _is_valid_placement(self, item_to_place_pos, item_to_place_dims, container_dims, occupied_volumes):
        '''
        Checks if placing an item at a given position is valid.

        Args:
            item_to_place_pos (tuple): The (x, y, z) position of the new item.
            item_to_place_dims (tuple): The (length, width, height) of the new item.
            container_dims (tuple): The (length, width, height) of the container.
            occupied_volumes (list): A list of dictionaries, where each dictionary
                                     represents an item already in the container.

        Returns:
            bool: True if the placement is valid, False otherwise.
        '''
        # Check if the item is within the container's boundaries.
        if not self._is_within_container_bounds(item_to_place_pos, item_to_place_dims, container_dims):
            return False

        # Check for overlaps with already placed items.
        for placed_item in occupied_volumes:
            placed_item_pos = (placed_item['x'], placed_item['y'], placed_item['z'])
            placed_item_dims = (placed_item['length'], placed_item['width'], placed_item['height'])
            if self._check_overlap_3d(item_to_place_pos, item_to_place_dims, placed_item_pos, placed_item_dims):
                return False
        
        # If no boundary violations or overlaps are found, the placement is valid.
        return True
```
Now, your task is to implement a class named `Algorithm` that inherits from `BaseAlgorithm`.
First, describe your novel algorithm and main steps for the `place_item` method in one sentence. This description must be a single Python comment line, starting with '# ', and the sentence itself must be enclosed in curly braces. (e.g., `# {This describes the algorithm using specific terms.}`)
Next, implement the `Algorithm` class.
This class must contain a method named `place_item`.

**To improve modularity and allow for future targeted optimizations, break down the logic within `place_item` into several private helper methods within the `Algorithm` class (e.g., methods for selecting an item, finding valid positions, evaluating placements, selecting a new truck, etc.).**

**Do not contain definition of base class in the output.**

The `place_item` method should accept 3 input(s): 'unplaced_items', 'trucks_in_use', 'truck_type'. The method should return 6 output(s): 'truck_index', 'item_index', 'x', 'y', 'z', 'orientation'.

        `unplaced_items`: A list of dictionaries, each representing an **item type**.
            - 'item_id', 'length', 'width', 'height' (these are the base dimensions)
            - 'quantity': int, the number of items of this type still needing to be packed.

        `trucks_in_use`: A list of dictionaries, where each dictionary represents a container currently in use.
            - `'occupied_volumes'`: A list of **dictionaries**. Each dictionary represents a placed item and has the following keys:
                - `'item_type_id'` (str): The unique identifier for the *type* of the placed item.
                - `'x'` (float): The starting position along the container's length (x-axis).
                - `'y'` (float): The starting position along the container's width (y-axis).
                - `'z'` (float): The starting position along the container's height (z-axis).
                - `'length'` (float): The dimension of the item along the x-axis **in its placed orientation**.
                - `'width'` (float): The dimension of the item along the y-axis **in its placed orientation**.
                - `'height'` (float): The dimension of the item along the z-axis **in its placed orientation**.
        `truck_type`: A **single tuple** for the standard container dimensions: `(length, width, height)`.
            
        **Return Values:**
        `truck_index`: Index of the container in `trucks_in_use`. Use -1 for a new container.
        `item_index`: Index of the item **type** in `unplaced_items` to place.
        `x`, `y`, `z`: Position for the new item. If a new container is requested (i.e., truck_index is -1), this must be the position for the item within that new container.
        `orientation`: An integer from 0 to 5, specifying the item's orientation. Let the original dimensions from `unplaced_items` be (L, W, H). The placed dimensions (length, width, height) will be:
            - 0: (L, W, H)
            - 1: (L, H, W)
            - 2: (W, L, H)
            - 3: (W, H, L)
            - 4: (H, L, W)
            - 5: (H, W, L)
        
The solution must not contain any comments or print statements.
Do not give additional explanations beyond the one-sentence algorithm description and the class code. Remove all comments before final output
\end{myverbatim}


\section{Results for the Base Problem}
\label{app:detailed_results}
Table~\ref{tab:ivancic_results_eoh} and ~\ref{tab:hybird_results_containers} provides a detailed, instance-by-instance comparison for the base problem. For a comprehensive breakdown of the other human-designed baselines presented in Table~\ref{tab:base_problem_results}, we refer the reader to the original study by \citet{kurpel2020exact}.

\begin{table*}[h!]
\centering
\small
\caption{Detailed instance-by-instance results for the base problem. Bold values indicate a match with the optimal solution.}
\label{tab:ivancic_results_eoh}
\begin{tabular}{@{}ccccccccc@{}}
\toprule
& \multicolumn{3}{c}{EoH: Budget time: 10} & \multicolumn{3}{c}{EoH: Budget time: 60} & & \\
\cmidrule(r){2-4} \cmidrule(r){5-7}
\textbf{\#} & \textbf{Seed: 0} & \textbf{Seed: 1} & \textbf{Seed: 2} & \textbf{Seed: 0} & \textbf{Seed: 1} & \textbf{Seed: 2} & \textbf{ZHU} & \textbf{Exact Solver} \\ \midrule
1 & \textbf{25} & \textbf{25} & 26 & \textbf{25} & \textbf{25} & \textbf{25} & \textbf{25} & \textbf{25} \\
2 & 10 & 10 & 10 & 10 & 10 & 10 & 10 & \textbf{9} \\
3 & 23 & 24 & 22 & \textbf{19} & 21 & 25 & \textbf{19} & \textbf{19} \\
4 & 28 & 28 & 31 & 27 & 30 & 28 & \textbf{26} & \textbf{26} \\
5 & 53 & 52 & 52 & \textbf{51} & 53 & 52 & \textbf{51} & \textbf{51} \\
6 & \textbf{10} & \textbf{10} & \textbf{10} & \textbf{10} & 11 & \textbf{10} & \textbf{10} & \textbf{10} \\
7 & \textbf{16} & \textbf{16} & \textbf{16} & \textbf{16} & \textbf{16} & \textbf{16} & \textbf{16} & \textbf{16} \\
8 & \textbf{4} & \textbf{4} & 5 & \textbf{4} & \textbf{4} & \textbf{4} & \textbf{4} & \textbf{4} \\
9 & 22 & 22 & \textbf{19} & \textbf{19} & 20 & 22 & \textbf{19} & \textbf{19} \\
10 & \textbf{55} & \textbf{55} & \textbf{55} & \textbf{55} & \textbf{55} & \textbf{55} & \textbf{55} & \textbf{55} \\
11 & 18 & 18 & 22 & 19 & \textbf{16} & 18 & \textbf{16} & \textbf{16} \\
12 & 56 & 56 & 56 & \textbf{53} & 56 & 56 & \textbf{53} & \textbf{53} \\
13 & 27 & 27 & 27 & 40 & 40 & 27 & \textbf{25} & \textbf{25} \\
14 & \textbf{27} & \textbf{27} & 28 & 28 & \textbf{27} & \textbf{27} & \textbf{27} & \textbf{27} \\
15 & 15 & 14 & 15 & 12 & 13 & 14 & \textbf{11} & \textbf{11} \\
16 & 33 & 33 & 34 & 32 & 32 & 33 & \textbf{26} & \textbf{26} \\
17 & 10 & 10 & 9 & 8 & 10 & 10 & \textbf{7} & \textbf{7} \\
18 & \textbf{2} & \textbf{2} & \textbf{2} & 3 & 3 & \textbf{2} & \textbf{2} & \textbf{2} \\
19 & \textbf{3} & \textbf{3} & 4 & \textbf{3} & 4 & \textbf{3} & \textbf{3} & \textbf{3} \\
20 & \textbf{5} & \textbf{5} & \textbf{5} & 6 & \textbf{5} & \textbf{5} & \textbf{5} & \textbf{5} \\
21 & 26 & 26 & 25 & 22 & 22 & 26 & \textbf{20} & \textbf{20} \\
22 & 10 & 11 & 11 & 10 & 10 & 10 & \textbf{8} & \textbf{8} \\
23 & 21 & 21 & 21 & 22 & 21 & 21 & \textbf{19} & \textbf{19} \\
24 & 7 & 7 & 8 & 6 & 7 & 7 & \textbf{5} & \textbf{5} \\
25 & \textbf{5} & 6 & \textbf{5} & 6 & 7 & 6 & \textbf{5} & \textbf{5} \\
26 & \textbf{3} & 4 & 4 & 4 & 9 & 4 & \textbf{3} & \textbf{3} \\
27 & 5 & 5 & 6 & 5 & 5 & 5 & \textbf{4} & \textbf{4} \\
28 & 12 & 12 & 12 & 10 & 11 & 12 & \textbf{9} & \textbf{9} \\
29 & 21 & 21 & 21 & 19 & 20 & 21 & 17 & \textbf{16} \\
30 & 28 & 28 & 27 & 30 & 26 & 28 & \textbf{22} & \textbf{22} \\
31 & 14 & 14 & 16 & 15 & 15 & 14 & \textbf{12} & 13 \\
32 & \textbf{4} & \textbf{4} & \textbf{4} & 6 & 5 & \textbf{4} & \textbf{4} & \textbf{4} \\
33 & 5 & 5 & 5 & 5 & 5 & 5 & \textbf{4} & \textbf{4} \\
34 & 11 & 11 & 9 & 9 & 9 & 11 & \textbf{8} & \textbf{8} \\
35 & 3 & 3 & 3 & 3 & 7 & 3 & \textbf{2} & \textbf{2} \\
36 & \textbf{14} & \textbf{14} & \textbf{14} & 17 & \textbf{14} & \textbf{14} & \textbf{14} & \textbf{14} \\
37 & \textbf{23} & \textbf{23} & \textbf{23} & \textbf{23} & 24 & \textbf{23} & \textbf{23} & \textbf{23} \\
38 & \textbf{45} & \textbf{45} & \textbf{45} & \textbf{45} & \textbf{45} & \textbf{45} & \textbf{45} & \textbf{45} \\
39 & 18 & 18 & 18 & 16 & 16 & 18 & \textbf{15} & \textbf{15} \\
40 & 12 & 12 & 10 & 10 & 10 & 12 & \textbf{8} & \textbf{8} \\
41 & 23 & 23 & 23 & 17 & 23 & 23 & \textbf{15} & \textbf{15} \\
42 & 5 & 5 & 7 & 5 & 6 & 5 & \textbf{4} & \textbf{4} \\
43 & 4 & 4 & 4 & 4 & 6 & 4 & \textbf{3} & \textbf{3} \\
44 & 4 & 4 & 4 & 4 & 9 & 4 & \textbf{3} & \textbf{3} \\
45 & \textbf{3} & \textbf{3} & \textbf{3} & \textbf{3} & 6 & \textbf{3} & \textbf{3} & \textbf{3} \\
46 & \textbf{2} & \textbf{2} & \textbf{2} & 3 & 8 & \textbf{2} & \textbf{2} & \textbf{2} \\
47 & 4 & 4 & 4 & 4 & 7 & 4 & \textbf{3} & \textbf{3} \\ \bottomrule
\end{tabular}
\end{table*}

\begin{table}[h!]
\centering
\caption{Detailed instance-by-instance results for the base problem with hybrid methods. Bold values indicate a match with the optimal solution.}
\label{tab:hybird_results_containers}

\begin{tabular}{@{}cccccc@{}}
\toprule
\textbf{\#} & \textbf{ZHU} & \textbf{Random best} & \textbf{S\_grasp best} & \textbf{LLM best} & \textbf{Exact Solver} \\ \midrule
1           & \textbf{25}  & \textbf{25}          & \textbf{25}            & \textbf{25}       & \textbf{25}         \\
2           & 10           & 10                   & 10                     & 10                & \textbf{9}          \\
3           & \textbf{19}  & \textbf{19}          & \textbf{19}            & \textbf{19}       & \textbf{19}         \\
4           & \textbf{26}  & \textbf{26}          & \textbf{26}            & \textbf{26}       & \textbf{26}         \\
5           & \textbf{51}  & \textbf{51}          & \textbf{51}            & \textbf{51}       & \textbf{51}         \\
6           & \textbf{10}  & \textbf{10}          & \textbf{10}            & \textbf{10}       & \textbf{10}         \\
7           & \textbf{16}  & \textbf{16}          & \textbf{16}            & \textbf{16}       & \textbf{16}         \\
8           & \textbf{4}   & 5                    & \textbf{4}             & \textbf{4}        & \textbf{4}          \\
9           & \textbf{19}  & \textbf{19}          & \textbf{19}            & \textbf{19}       & \textbf{19}         \\
10          & \textbf{55}  & \textbf{55}          & \textbf{55}            & \textbf{55}       & \textbf{55}         \\
11          & \textbf{16}  & \textbf{16}          & 19                     & \textbf{16}       & \textbf{16}         \\
12          & \textbf{53}  & \textbf{53}          & \textbf{53}            & \textbf{53}       & \textbf{53}         \\
13          & \textbf{25}  & \textbf{25}          & \textbf{25}            & \textbf{25}       & \textbf{25}         \\
14          & \textbf{27}  & \textbf{27}          & \textbf{27}            & \textbf{27}       & \textbf{27}         \\
15          & \textbf{11}  & \textbf{11}          & \textbf{11}            & \textbf{11}       & \textbf{11}         \\
16          & \textbf{26}  & \textbf{26}          & \textbf{26}            & \textbf{26}       & \textbf{26}         \\
17          & \textbf{7}   & 8                    & \textbf{7}             & \textbf{7}        & \textbf{7}          \\
18          & \textbf{2}   & 3                    & \textbf{2}             & \textbf{2}        & \textbf{2}          \\
19          & \textbf{3}   & 4                    & \textbf{3}             & \textbf{3}        & \textbf{3}          \\
20          & \textbf{5}   & \textbf{5}           & \textbf{5}             & \textbf{5}        & \textbf{5}          \\
21          & \textbf{20}  & \textbf{20}          & \textbf{20}            & \textbf{20}       & \textbf{20}         \\
22          & \textbf{8}   & 10                   & 9                      & \textbf{8}        & \textbf{8}          \\
23          & \textbf{19}  & 20                   & 20                     & \textbf{19}       & \textbf{19}         \\
24          & \textbf{5}   & 6                    & \textbf{5}             & \textbf{5}        & \textbf{5}          \\
25          & \textbf{5}   & 6                    & \textbf{5}             & \textbf{5}        & \textbf{5}          \\
26          & \textbf{3}   & 4                    & \textbf{3}             & \textbf{3}        & \textbf{3}          \\
27          & \textbf{4}   & 6                    & 5                      & 5                 & \textbf{4}          \\
28          & \textbf{9}   & 10                   & 10                     & \textbf{9}        & \textbf{9}          \\
29          & 17           & 17                   & 17                     & 17                & \textbf{16}         \\
30          & \textbf{22}  & \textbf{22}          & 23                     & \textbf{22}       & \textbf{22}         \\
31          & \textbf{12}  & 13                   & \textbf{12}            & \textbf{12}       & 13                  \\
32          & \textbf{4}   & \textbf{4}           & \textbf{4}             & \textbf{4}        & \textbf{4}          \\
33          & \textbf{4}   & 5                    & 5                      & 5                 & \textbf{4}          \\
34          & \textbf{8}   & 9                    & \textbf{8}             & \textbf{8}        & \textbf{8}          \\
35          & \textbf{2}   & 3                    & \textbf{2}             & 3                 & \textbf{2}          \\
36          & \textbf{14}  & \textbf{14}          & \textbf{14}            & \textbf{14}       & \textbf{14}         \\
37          & \textbf{23}  & \textbf{23}          & \textbf{23}            & \textbf{23}       & \textbf{23}         \\
38          & \textbf{45}  & \textbf{45}          & \textbf{45}            & \textbf{45}       & \textbf{45}         \\
39          & \textbf{15}  & \textbf{15}          & \textbf{15}            & \textbf{15}       & \textbf{15}         \\
40          & \textbf{8}   & 9                    & \textbf{8}             & \textbf{8}        & \textbf{8}          \\
41          & \textbf{15}  & \textbf{15}          & \textbf{15}            & \textbf{15}       & \textbf{15}         \\
42          & \textbf{4}   & 5                    & \textbf{4}             & \textbf{4}        & \textbf{4}          \\
43          & \textbf{3}   & 4                    & \textbf{3}             & \textbf{3}        & \textbf{3}          \\
44          & \textbf{3}   & 4                    & \textbf{3}             & 4                 & \textbf{3}          \\
45          & \textbf{3}   & \textbf{3}           & \textbf{3}             & \textbf{3}        & \textbf{3}          \\
46          & \textbf{2}   & 3                    & \textbf{2}             & \textbf{2}        & \textbf{2}          \\
47          & \textbf{3}   & 4                    & \textbf{3}             & \textbf{3}        & \textbf{3}          \\ \bottomrule
\end{tabular}
\end{table}

\FloatBarrier 

\section{Results for the Constrained Problem}
\label{app:detailed_results_constrained}
Table~\ref{tab:stability_results},~\ref{tab:comparison_separation_boxes} and \ref{tab:comparison_results} provide a detailed, instance-by-instance comparison for the constrained problem. 

\begin{table}[htbp]
\centering
\caption{Detailed instance-by-instance results for the problem with load stability. Bold values indicate a match with the optimal solution.}
\label{tab:stability_results}
\begin{tabular}{@{}rrrrrrrrr@{}}
\toprule
\multicolumn{1}{c}{\#} & \multicolumn{1}{c}{ELS} & \multicolumn{1}{c}{ELP} & \multicolumn{1}{c}{ZST} & \multicolumn{1}{c}{Exact Solver} & \multicolumn{1}{c}{EoH} & \multicolumn{1}{c}{EoH} & \multicolumn{1}{c}{EoH} & \multicolumn{1}{c}{hybrid} \\
& & & & \multicolumn{1}{c}{} & \multicolumn{1}{c}{Seed: 0} & \multicolumn{1}{c}{Seed: 1} & \multicolumn{1}{c}{Seed: 2} & \multicolumn{1}{c}{best} \\
\midrule
1  & 27 & 26 & \textbf{25} & \textbf{25} & \textbf{25} & \textbf{25} & \textbf{25} & \textbf{25} \\
2  & 11 & \textbf{10} & \textbf{10} & \textbf{10} & \textbf{10} & \textbf{10} & \textbf{10} & \textbf{10} \\
3  & 21 & 22 & \textbf{19} & \textbf{19} & 22 & 23 & 22 & \textbf{19} \\
4  & 29 & 30 & \textbf{26} & \textbf{26} & 27 & \textbf{26} & 28 & \textbf{26} \\
5  & 55 & \textbf{51} & \textbf{51} & \textbf{51} & \textbf{51} & \textbf{51} & 52 & \textbf{51} \\
6  & \textbf{10} & \textbf{10} & \textbf{10} & \textbf{10} & \textbf{10} & \textbf{10} & \textbf{10} & \textbf{10} \\
7  & \textbf{16} & \textbf{16} & \textbf{16} & \textbf{16} & \textbf{16} & \textbf{16} & \textbf{16} & \textbf{16} \\
8  & \textbf{4}  & \textbf{4}  & \textbf{4}  & \textbf{4} & 5 & \textbf{4} & \textbf{4} & \textbf{4} \\
9  & \textbf{19} & \textbf{19} & \textbf{19} & \textbf{19} & 22 & 22 & 22 & \textbf{19} \\
10 & \textbf{55} & \textbf{55} & \textbf{55} & \textbf{55} & \textbf{55} & \textbf{55} & \textbf{55} & \textbf{55} \\
11 & 18 & 17 & \textbf{16} & \textbf{16} & 18 & 18 & 18 & \textbf{16} \\
12 & \textbf{53} & \textbf{53} & \textbf{53} & \textbf{53} & 56 & 56 & 56 & \textbf{53} \\
13 & \textbf{25} & \textbf{25} & \textbf{25} & \textbf{25} & 28 & 27 & 27 & \textbf{25} \\
14 & 28 & \textbf{27} & \textbf{27} & \textbf{27} & 28 & \textbf{27} & \textbf{27} & \textbf{27} \\
15 & 12 & 12 & \textbf{11} & \textbf{11} & 16 & 15 & 14 & \textbf{11} \\
16 & 28 & \textbf{26} & \textbf{26} & \textbf{26} & 33 & 33 & 33 & \textbf{26} \\
17 & 8  & \textbf{7} & \textbf{7} & \textbf{7} & 9 & 10 & 10 & 8 \\
18 & \textbf{2}  & \textbf{2}  & \textbf{2}  & \textbf{2} & 3 & \textbf{2} & \textbf{2} & \textbf{2} \\
19 & \textbf{3}  & \textbf{3}  & \textbf{3}  & \textbf{3} & 4 & 4 & \textbf{3} & \textbf{3} \\
20 & \textbf{5}  & \textbf{5}  & \textbf{5}  & \textbf{5} & \textbf{5} & \textbf{5} & \textbf{5} & \textbf{5} \\
21 & 24 & 26 & \textbf{20} & \textbf{20} & 27 & 26 & 26 & \textbf{20} \\
22 & 9  & \textbf{8} & \textbf{8} & \textbf{8} & 9 & 9 & 10 & 9 \\
23 & 21 & \textbf{20} & 21 & 21 & 22 & 21 & 22 & \textbf{20} \\
24 & 6  & \textbf{5} & \textbf{5} & \textbf{5} & 7 & 7 & 8 & 6 \\
25 & 6  & \textbf{5} & \textbf{5} & \textbf{5} & 6 & 6 & \textbf{5} & \textbf{5} \\
26 & \textbf{3}  & \textbf{3}  & \textbf{3}  & \textbf{3} & 4 & \textbf{3} & 4 & \textbf{3} \\
27 & 5  & 5  & \textbf{4} & \textbf{4} & 5 & 5 & 5 & 5 \\
28 & 11 & 10 & 10 & \textbf{9} & 11 & 12 & 12 & 10 \\
29 & 18 & \textbf{17} & \textbf{17} & \textbf{17} & 22 & 21 & 21 & \textbf{17} \\
30 & \textbf{22} & \textbf{22} & \textbf{22} & \textbf{22} & 26 & 28 & 28 & \textbf{22} \\
31 & 13 & \textbf{12} & 13 & 13 & 14 & 15 & 14 & \textbf{12} \\
32 & \textbf{4}  & \textbf{4}  & \textbf{4}  & \textbf{4} & \textbf{4} & \textbf{4} & \textbf{4} & \textbf{4} \\
33 & 5  & \textbf{4} & \textbf{4} & \textbf{4} & 5 & 5 & 5 & 5 \\
34 & \textbf{8}  & \textbf{8}  & \textbf{8}  & \textbf{8} & 11 & 11 & 11 & \textbf{8} \\
35 & \textbf{2}  & \textbf{2}  & \textbf{2}  & \textbf{2} & 3 & 3 & 3 & 3 \\
36 & 18 & \textbf{14} & \textbf{14} & \textbf{14} & \textbf{14} & \textbf{14} & \textbf{14} & \textbf{14} \\
37 & 26 & \textbf{23} & \textbf{23} & \textbf{23} & \textbf{23} & \textbf{23} & \textbf{23} & \textbf{23} \\
38 & 46 & \textbf{45} & \textbf{45} & \textbf{45} & \textbf{45} & \textbf{45} & \textbf{45} & \textbf{45} \\
39 & \textbf{15} & \textbf{15} & \textbf{15} & \textbf{15} & 19 & 18 & 18 & \textbf{15} \\
40 & 9  & 9  & \textbf{8} & \textbf{8} & 12 & 12 & 12 & \textbf{8} \\
41 & 16 & \textbf{15} & \textbf{15} & \textbf{15} & 23 & 23 & 23 & \textbf{15} \\
42 & \textbf{4}  & \textbf{4}  & \textbf{4}  & \textbf{4} & 5 & 5 & 5 & \textbf{4} \\
43 & \textbf{3}  & \textbf{3}  & \textbf{3}  & \textbf{3} & 4 & 4 & 4 & \textbf{3} \\
44 & 4  & \textbf{3} & \textbf{3} & \textbf{3} & 5 & 4 & 4 & 4 \\
45 & \textbf{3}  & \textbf{3}  & \textbf{3}  & \textbf{3} & \textbf{3} & \textbf{3} & \textbf{3} & \textbf{3} \\
46 & \textbf{2}  & \textbf{2}  & \textbf{2}  & \textbf{2} & \textbf{2} & \textbf{2} & \textbf{2} & \textbf{2} \\
47 & \textbf{3}  & \textbf{3}  & \textbf{3}  & \textbf{3} & 4 & 4 & 4 & 4 \\
\bottomrule
\end{tabular}
\end{table}

\begin{table}[h!]
\centering
\caption{Detailed instance-by-instance results for the problem with separation of boxes. Bold values indicate a match with the optimal solution.}
\label{tab:comparison_separation_boxes}
\begin{tabular}{ccccccc}
\toprule
\# & ELY & \begin{tabular}[c]{@{}c@{}}Exact Solver\end{tabular} & \begin{tabular}[c]{@{}c@{}}EoH\\ seed 0\end{tabular} & \begin{tabular}[c]{@{}c@{}}EoH\\ seed 1\end{tabular} & \begin{tabular}[c]{@{}c@{}}EoH\\ seed 2\end{tabular} & \begin{tabular}[c]{@{}c@{}}hybrid\\ best\end{tabular} \\
\midrule
1  & \textbf{27} & \textbf{27} & \textbf{27} & \textbf{27} & \textbf{27} & \textbf{27} \\
2  & 11 & \textbf{10} & 11 & 11 & 11 & 11 \\
3  & \textbf{20} & \textbf{20} & 23 & 23 & 24 & 21 \\
4  & \textbf{28} & \textbf{28} & 30 & 31 & 29 & 29 \\
5  & \textbf{51} & \textbf{51} & 52 & 56 & 59 & 52 \\
6  & \textbf{10} & \textbf{10} & 11 & 11 & 11 & 11 \\
7  & \textbf{16} & \textbf{16} & \textbf{16} & \textbf{16} & \textbf{16} & \textbf{16} \\
8  & \textbf{4}  & \textbf{4}  & 5  & 5  & 5  & 5  \\
9  & \textbf{19} & \textbf{19} & 22 & 21 & 22 & \textbf{19} \\
10 & \textbf{55} & \textbf{55} & \textbf{55} & \textbf{55} & \textbf{55} & \textbf{55} \\
11 & 18 & \textbf{17} & 18 & 25 & 18 & 19 \\
12 & \textbf{56} & \textbf{56} & \textbf{56} & 57 & \textbf{56} & \textbf{56} \\
13 & \textbf{25} & \textbf{25} & 27 & 27 & 27 & \textbf{25} \\
14 & \textbf{28} & \textbf{28} & 29 & 29 & 29 & \textbf{28} \\
15 & \textbf{12} & \textbf{12} & 15 & 15 & 15 & \textbf{12} \\
16 & \textbf{26} & \textbf{26} & 33 & 33 & 33 & \textbf{26} \\
17 & \textbf{9}  & \textbf{9}  & 11 & 11 & 11 & \textbf{9}  \\
18 & \textbf{2}  & \textbf{2}  & 3  & 3  & 3  & \textbf{2}  \\
19 & \textbf{3}  & \textbf{3}  & \textbf{3}  & 4  & 4  & \textbf{3}  \\
20 & \textbf{5}  & \textbf{5}  & 7  & 6  & 6  & \textbf{5}  \\
21 & 21 & \textbf{20} & 26 & 26 & 26 & 21 \\
22 & \textbf{8}  & \textbf{8}  & 12 & 11 & 10 & 9  \\
23 & 21 & \textbf{20} & 23 & 21 & 21 & \textbf{20} \\
24 & 6  & \textbf{5}  & 8  & 8  & 8  & 6  \\
25 & \textbf{5}  & \textbf{5}  & 6  & 6  & 6  & 6  \\
26 & \textbf{3}  & \textbf{3}  & 4  & 4  & 4  & 4  \\
27 & \textbf{5}  & \textbf{5}  & \textbf{5}  & \textbf{5}  & \textbf{5}  & \textbf{5}  \\
28 & 10 & \textbf{9}  & 11 & 11 & 12 & 10 \\
29 & 17 & \textbf{16} & 24 & 22 & 22 & 17 \\
30 & 23 & \textbf{22} & 28 & 26 & 31 & 25 \\
31 & \textbf{13} & \textbf{13} & 16 & 15 & 15 & \textbf{13} \\
32 & \textbf{4}  & \textbf{4}  & 5  & 5  & 5  & \textbf{4}  \\
33 & \textbf{5}  & \textbf{5}  & \textbf{5}  & \textbf{5}  & 6  & \textbf{5}  \\
34 & 9  & \textbf{8}  & 11 & 9  & 10 & 9  \\
35 & \textbf{3}  & \textbf{3}  & \textbf{3}  & \textbf{3}  & \textbf{3}  & \textbf{3}  \\
36 & \textbf{18} & \textbf{18} & 19 & 19 & 19 & 19 \\
37 & \textbf{23} & \textbf{23} & \textbf{23} & \textbf{23} & \textbf{23} & \textbf{23} \\
38 & \textbf{45} & \textbf{45} & \textbf{45} & \textbf{45} & \textbf{45} & \textbf{45} \\
39 & \textbf{15} & \textbf{15} & 19 & 19 & 19 & \textbf{15} \\
40 & \textbf{9}  & \textbf{9}  & 12 & 10 & 12 & \textbf{9}  \\
41 & \textbf{16} & \textbf{16} & 23 & 23 & 23 & 17 \\
42 & 5  & \textbf{4}  & 5  & 7  & 5  & 5  \\
43 & 4  & \textbf{3}  & 5  & 4  & 4  & \textbf{3}  \\
44 & \textbf{4}  & \textbf{4}  & \textbf{4}  & \textbf{4}  & \textbf{4}  & \textbf{4}  \\
45 & \textbf{3}  & \textbf{3}  & \textbf{3}  & \textbf{3}  & \textbf{3}  & \textbf{3}  \\
46 & \textbf{2}  & \textbf{2}  & \textbf{2}  & \textbf{2}  & \textbf{2}  & \textbf{2}  \\
47 & 3  & \textbf{4}  & 5  & \textbf{4}  & 5  & \textbf{4}  \\
\bottomrule
\end{tabular}
\end{table}

\begin{table}[h]
\centering
\caption{Detailed instance-by-instance results for the problem with stability and separation constraints. Bold values indicate a match with the optimal solution.}
\label{tab:comparison_results}
\begin{tabular}{@{}cccccc@{}}
\toprule
\# & \begin{tabular}[c]{@{}c@{}}Exact Solver\end{tabular} & \begin{tabular}[c]{@{}c@{}}EoH\\ seed 0\end{tabular} & \begin{tabular}[c]{@{}c@{}}EoH\\ seed 1\end{tabular} & \begin{tabular}[c]{@{}c@{}}EoH\\ seed 2\end{tabular} & \begin{tabular}[c]{@{}c@{}}hybrid\\ best\end{tabular} \\ \midrule
1  & \textbf{27} & \textbf{27} & \textbf{27} & \textbf{27} & \textbf{27} \\
2  & \textbf{10} & 11 & 11 & 11 & 11 \\
3  & \textbf{20} & 27 & 25 & 25 & 22 \\
4  & \textbf{28} & 29 & \textbf{28} & \textbf{28} & 29 \\
5  & \textbf{51} & 59 & 52 & 52 & \textbf{51} \\
6  & \textbf{10} & 11 & 11 & 11 & \textbf{10} \\
7  & \textbf{16} & \textbf{16} & 17 & \textbf{16} & \textbf{16} \\
8  & \textbf{4}  & 5  & 5  & 5  & \textbf{4}  \\
9  & \textbf{19} & 22 & 22 & 22 & 21 \\
10 & \textbf{55} & \textbf{55} & \textbf{55} & \textbf{55} & \textbf{55} \\
11 & 18 & 18 & 18 & 18 & \textbf{17} \\
12 & \textbf{56} & \textbf{56} & \textbf{56} & \textbf{56} & \textbf{56} \\
13 & \textbf{25} & 27 & 27 & 27 & \textbf{25} \\
14 & \textbf{28} & 29 & 29 & 29 & 29 \\
15 & \textbf{12} & 15 & 15 & 15 & 13 \\
16 & \textbf{26} & 33 & 33 & 33 & 29 \\
17 & \textbf{9}  & 11 & 11 & 11 & 10 \\
18 & \textbf{2}  & 3  & 3  & 3  & \textbf{2}  \\
19 & \textbf{3}  & 6  & 4  & 4  & \textbf{3}  \\
20 & \textbf{5}  & 7  & 6  & 6  & \textbf{5}  \\
21 & \textbf{20} & 31 & 27 & 27 & 23 \\
22 & \textbf{8}  & 17 & 11 & 12 & 9  \\
23 & 21 & 33 & 21 & 22 & \textbf{20} \\
24 & \textbf{5}  & 9  & 7  & 8  & 6  \\
25 & \textbf{5}  & 8  & 6  & 6  & \textbf{5}  \\
26 & \textbf{3}  & 5  & 4  & 4  & 4  \\
27 & \textbf{5}  & 8  & 6  & \textbf{5}  & \textbf{5}  \\
28 & \textbf{10} & 13 & 12 & 12 & 11 \\
29 & \textbf{17} & 23 & 21 & 20 & 18 \\
30 & \textbf{23} & 37 & 31 & 31 & 25 \\
31 & \textbf{13} & 17 & 15 & 14 & \textbf{13} \\
32 & \textbf{4}  & 5  & \textbf{4}  & 5  & \textbf{4}  \\
33 & \textbf{5}  & 6  & 6  & 6  & \textbf{5}  \\
34 & \textbf{8}  & 11 & 11 & 10 & 9  \\
35 & \textbf{3}  & \textbf{3}  & \textbf{3}  & \textbf{3}  & \textbf{3}  \\
36 & \textbf{18} & 19 & 19 & 19 & \textbf{18} \\
37 & \textbf{23} & 27 & \textbf{23} & \textbf{23} & \textbf{23} \\
38 & \textbf{45} & 58 & \textbf{45} & \textbf{45} & \textbf{45} \\
39 & \textbf{15} & 21 & 19 & 19 & 17 \\
40 & \textbf{9}  & 13 & 12 & 12 & \textbf{9}  \\
41 & \textbf{16} & 25 & 23 & 23 & 18 \\
42 & \textbf{4}  & 5  & 5  & 5  & 5  \\
43 & \textbf{3}  & 5  & 5  & 4  & \textbf{3}  \\
44 & \textbf{4}  & 6  & 5  & \textbf{4}  & \textbf{4}  \\
45 & \textbf{3}  & 4  & \textbf{3}  & \textbf{3}  & \textbf{3}  \\
46 & \textbf{2}  & 3  & \textbf{2}  & \textbf{2}  & \textbf{2}  \\
47 & \textbf{4}  & 5  & 5  & 5  & \textbf{4}  \\ \bottomrule
\end{tabular}
\vspace{2mm} 
\par
\end{table}

\FloatBarrier 

\section{Benchmark Method Descriptions}
\label{app:benchmark_details}
This section provides details on the human-designed benchmark methods referenced.

\textbf{First-Fit.} A simple, deterministic greedy method that establishes a basic performance benchmark. Items are pre-sorted by volume in descending order. The algorithm generates candidate placement points at the container's origin and at the corners of previously packed items. It then iterates through these points—ordered by their x, y, and then z coordinates—and places the current item at the first valid location where it fits.

\textbf{IVA \citep{ivancic1988integer}.} This method uses an iterative heuristic guided by an overall integer programming (IP) formulation to build the solution one container at a time. In each iteration, a packing heuristic is run for each available container type to find the best packing pattern for the remaining items. These patterns are evaluated using a selection criterion from the IP model's objective function. The container offering the "best value" is selected, its contents are permanently assigned, and the process repeats until all items are packed.

\textbf{BOR \citep{bortfeldt2000heuristik}.} This paper proposes a metaheuristic approach that improves upon the standard sequential strategy of filling containers one by one. The method first constructs an initial solution sequentially and then employs a search algorithm to iteratively refine and enhance that solution.

\textbf{ELS \citep{eley2002solving} \& ELP \citep{eley2002solving}.} These tree search heuristics, presented in the paper for both single and multiple container problems, aim to find stable loading patterns and optimize volume utilization. ELS (tree search with sequential packing) fills containers one by one, whereas ELP (tree search with parallel packing) considers multiple containers simultaneously. Both methods systematically explore item loading sequences and orientations, building homogeneous blocks of identical items and evaluating partial solutions using a greedy heuristic to determine the potential for filling remaining space.

\textbf{ELY \citep{eley2003bottleneck}.} This work presents a two-stage "bottleneck assignment" approach based on a \textbf{set partitioning formulation}. In the first stage, a single-container packing heuristic generates a large set of high-quality, diverse packing patterns (columns). In the second stage, an integer programming solver addresses a set partitioning problem to select the optimal combination of these pre-generated patterns, aiming to pack all items while minimizing cost or the number of containers.

\textbf{LZG \citep{lim2005container}.} This work proposes a two-part heuristic that fills containers sequentially. The inner loop uses a layer-based packing algorithm, which identifies the lowest available surface in a container and solves a 2D packing problem to create a stable layer. The outer loop employs an iterative refinement scheme: it packs all items, analyzes any packing failures, adjusts item priorities accordingly, and then re-initiates the entire packing process from scratch.


\textbf{CHE \citep{che2011multiple} \& ZHU\citep{zhu2012prototype} \& ZST \citep{zhu2012prototype}.} These methods are advanced heuristics based on \textbf{column generation}. Unlike item-by-item placement, this two-stage approach first uses a "subproblem" heuristic to generate a set of complete, high-quality packing patterns (columns) for a single container. Then, a "master problem," typically a set covering formulation, selects the optimal combination of these patterns to pack all items using the minimum number of containers. The work by Zhu et al. extends this framework by incorporating practical constraints such as cargo stability.

\textbf{S-GRASP \citep{che2011multiple}.} This method is a sequential, constructive heuristic based on the Greedy Randomized Adaptive Search Procedure (GRASP) framework. It fills one container at a time by iteratively placing "walls" of identical items. The key innovation is a sophisticated look-ahead evaluation function used in the greedy selection step. The heuristic evaluates potential placements not just on their immediate volume utilization, but on the quality of the "residual spaces" they leave behind. It explicitly penalizes choices that fragment the container into small, unusable areas, thereby preserving larger, more valuable spaces for subsequent items. By randomly selecting from a Restricted Candidate List (RCL) of these high-quality options, the algorithm explores multiple promising solution paths to avoid premature convergence to a poor local optimum.

\textbf{Exact Solver \citep{kurpel2020exact}.} This method provides the ground truth for solution quality by formulating the problem as a 0-1 Integer Linear Program (ILP) and solving it with a state-of-the-art commercial solver to find a mathematically proven optimal solution. The primary contribution of this work is the development of advanced discretization techniques that reduce the number of potential placement points, making the ILP model tractable for benchmark instances that were previously unsolvable by exact methods.

\section{Examples of Evolved Scoring Functions}
\label{app:scoring_functions}
This section presents concrete examples of the scoring functions discovered through the Evolution of Heuristics (EoH) process, as discussed in Section~\ref{sec:experiments}. The code is presented as generated, including the LLM's own comments. 

The example below is one of the high-performing heuristics from our experiments. A more comprehensive collection of all discovered heuristics, showcasing the full evolutionary trajectory and diversity of solutions, is available in our public code repository.\footnote{\url{https://github.com/quangr/EoH}}

\begin{myverbatim}
class Algorithm(BaseAlgorithm):
    # {This algorithm prioritizes filling trucks completely by maximizing volume utilization and reducing the influence of item quantity and adjacency on the placement score.}
    def __init__(self, epsilon=1e-6):
        super().__init__(epsilon)

    def place_item(self, unplaced_items, trucks_in_use, truck_type):
        best_placement = self._find_best_placement(unplaced_items, trucks_in_use, truck_type)

        if best_placement:
            return best_placement['truck_index'], best_placement['item_index'], best_placement['x'], best_placement['y'], best_placement['z'], best_placement['orientation']
        else:
            return -1, -1, 0.0, 0.0, 0.0, 0

    def _find_best_placement(self, unplaced_items, trucks_in_use, truck_type):
        best_placement = None
        best_score = -1.0

        for item_index, item in enumerate(unplaced_items):
            if item['quantity'] > 0:
                for truck_index, truck in enumerate(trucks_in_use):
                    placement = self._find_placement_in_truck(item, item_index, truck, truck_index, truck_type)
                    if placement and placement['score'] > best_score:
                        best_score = placement['score']
                        best_placement = placement

                new_truck_placement = self._find_placement_in_new_truck(item, item_index, truck_type)
                if new_truck_placement and new_truck_placement['score'] > best_score:
                    best_score = new_truck_placement['score']
                    best_placement = new_truck_placement

        return best_placement

    def _find_placement_in_truck(self, item, item_index, truck, truck_index, truck_type):
        best_placement = None
        best_score = -1.0

        for orientation_index, (orientation_id, (l, w, h)) in enumerate(self._get_orientations(item)):
            potential_positions = self._generate_potential_positions(truck, truck_type, l, w, h)
            for px, py, pz in potential_positions:
                if self._is_valid_placement((px, py, pz), (l, w, h), truck_type, truck['occupied_volumes']):
                    score = self._calculate_placement_score((px, py, pz), (l, w, h), truck_type, truck['occupied_volumes'], item['quantity'])
                    if score > best_score:
                        best_score = score
                        best_placement = {
                            'truck_index': truck_index,
                            'item_index': item_index,
                            'x': px,
                            'y': py,
                            'z': pz,
                            'orientation': orientation_index,
                            'score': score
                        }
        return best_placement

    def _find_placement_in_new_truck(self, item, item_index, truck_type):
        best_placement = None
        best_score = -1.0
        empty_truck = {'occupied_volumes': []}

        for orientation_index, (orientation_id, (l, w, h)) in enumerate(self._get_orientations(item)):
            if self._is_valid_placement((0.0, 0.0, 0.0), (l, w, h), truck_type, empty_truck['occupied_volumes']):
                score = self._calculate_placement_score((0.0, 0.0, 0.0), (l, w, h), truck_type, empty_truck['occupied_volumes'], item['quantity'])
                if score > best_score:
                    best_score = score
                    best_placement = {
                        'truck_index': -1,
                        'item_index': item_index,
                        'x': 0.0,
                        'y': 0.0,
                        'z': 0.0,
                        'orientation': orientation_index,
                        'score': score
                    }
        return best_placement

    def _generate_potential_positions(self, truck, truck_type, l, w, h):
        potential_positions = []
        occupied_volumes = truck['occupied_volumes']
        cl, cw, ch = truck_type

        if not occupied_volumes:
            potential_positions.append((0.0, 0.0, 0.0))
            return potential_positions

        for placed_item in occupied_volumes:
            x = placed_item['x']
            y = placed_item['y']
            z = placed_item['z']
            length = placed_item['length']
            width = placed_item['width']
            height = placed_item['height']

            new_x = x + length
            new_y = y + width
            new_z = z + height

            if new_x + l <= cl + self.epsilon:
                potential_positions.append((new_x, y, z))
            if new_y + w <= cw + self.epsilon:
                potential_positions.append((x, new_y, z))
            if new_z + h <= ch + self.epsilon:
                potential_positions.append((x, y, new_z))

        potential_positions.append((0.0, 0.0, 0.0))
        
        return potential_positions

    def _calculate_placement_score(self, item_pos, item_dims, container_dims, occupied_volumes, item_quantity):
        volume_utilization = self._calculate_volume_utilization(item_dims, container_dims)
        item_priority = item_quantity
        adjacency_score = self._calculate_adjacency_score(item_pos, item_dims, container_dims, occupied_volumes)

        score = (0.9 * volume_utilization +
                 0.05 * item_priority +
                 0.05 * adjacency_score)

        return score

    def _calculate_volume_utilization(self, item_dims, container_dims):
        item_volume = item_dims[0] * item_dims[1] * item_dims[2]
        container_volume = container_dims[0] * container_dims[1] * container_dims[2]
        return item_volume / container_volume

    def _calculate_adjacency_score(self, item_pos, item_dims, container_dims, occupied_volumes):
        px, py, pz = item_pos
        l, w, h = item_dims
        cl, cw, ch = container_dims

        adjacency = 0

        if px == 0.0: adjacency += 1
        if py == 0.0: adjacency += 1
        if pz == 0.0: adjacency += 1
        if px + l >= cl - self.epsilon: adjacency += 1
        if py + w >= cw - self.epsilon: adjacency += 1
        if pz + h >= ch - self.epsilon: adjacency += 1

        for placed_item in occupied_volumes:
            x = placed_item['x']
            y = placed_item['y']
            z = placed_item['z']
            length = placed_item['length']
            width = placed_item['width']
            height = placed_item['height']

            if abs(px + l - x) < self.epsilon: adjacency += 1
            if abs(py + w - y) < self.epsilon: adjacency += 1
            if abs(pz + h - z) < self.epsilon: adjacency += 1
            if abs(x + length - px) < self.epsilon: adjacency += 1
            if abs(y + width - py) < self.epsilon: adjacency += 1
            if abs(z + height - pz) < self.epsilon: adjacency += 1

        return adjacency
\end{myverbatim}

The following are specific examples of the scoring criteria that emerged during the evolutionary process, as discussed in the main text.

\subsection{Evolved Item Selection Criteria}
The initial simple heuristic, which considered only an item's volume, underwent a significant evolution to incorporate more criteria. Examples include:

\begin{enumerate}
    \item \textbf{Large-and-Numerous Priority:} Items that are both large \emph{and} plentiful receive higher scores.
    \[
        \text{score} = \text{volume} \times \text{quantity}
    \]

    \item \textbf{Cubeness Measure:} This acts as a proxy for shape compactness, favoring near-cubic items that are generally easier to pack.
    \[
        \text{score} = \frac{\text{volume}}{\text{length} + \text{width} + \text{height}}
    \]

    \item \textbf{Space-Fitting Preference:} This prioritizes items that fit most efficiently into containers that are nearing capacity.
    \[
        \text{score} = \text{item\_volume} \times \left( 1 - \frac{\text{remaining\_volume}}{\text{total\_volume}} \right)
    \]
\end{enumerate}

\subsection{Evolved Placement Evaluation Logic}
The most significant evolution occurred in how a placement position is chosen, where the LLM invented and refined scoring functions to evaluate all valid candidate positions. Key strategies that emerged include:

\begin{itemize}
    \item \textbf{Minimizing Wasted Space:} Early versions simply tried to minimize the bounding box of the packed items. Later versions evolved more specific waste-calculation functions, such as computing the sum of leftover gaps along each axis.

    \item \textbf{Minimizing Height:} A common strategy that emerged was to favor placements that minimally increase the maximum $z$-coordinate (the ``height'' of the pack). By minimizing this, the heuristic learns to build level layers, a key principle in efficient packing.

    \item \textbf{Maximizing Contact Area:} Some heuristics learned to favor placements that maximize surface contact with other boxes or the container walls, leading to denser and more stable configurations.
\end{itemize}

\end{document}